\documentclass[10pt,twocolumn,letterpaper]{article}

\usepackage{cvpr}
\usepackage{times}
\usepackage{epsfig}
\usepackage{graphicx}
\usepackage{amsmath}
\usepackage{amssymb}

\usepackage[pagebackref=true,breaklinks=true,letterpaper=true,colorlinks,bookmarks=false]{hyperref}

\cvprfinalcopy % *** Uncomment this line for the final submission

\def\cvprPaperID{24} % *** Enter the CVPR Paper ID here

\ifcvprfinal\fi
\begin{document}

%%%%%%%%% TITLE
\title{Ultra Power-Efficient CNN Domain Specific Accelerator with 9.3TOPS/Watt for Mobile and Embedded Applications}

\author{Baohua Sun,  ~~~~Lin Yang, ~~~~Patrick Dong,  ~~~~Wenhan Zhang,  ~~~~Jason Dong,  ~~~~Charles Young\\
Gyrfalcon Technology Inc.\\
1900 McCarthy Blvd. Milpitas, CA 95035\\
{\tt\small \{baohua.sun,lin.yang,patrick.dong,wenhan.zhang,jason.dong,charles.yang\}@gyrfalcontech.com}
%\and
%Lin Yang\\
}

\maketitle
\thispagestyle{empty}

%%%%%%%%% ABSTRACT
\begin{abstract}
   Computer vision performances have been significantly improved in recent years by Convolutional Neural Networks (CNN). Currently, applications using CNN algorithms are deployed mainly on general purpose hardwares, such as CPUs, GPUs or FPGAs. However, power consumption, speed, accuracy, memory footprint, and die size should all be taken into consideration for mobile and embedded applications. Domain Specific Architecture (DSA) for CNN is the efficient and practical solution for CNN deployment and implementation. We designed and produced a 28nm Two-Dimensional CNN-DSA accelerator with an ultra power-efficient performance of 9.3TOPS/Watt and with all processing done in the internal memory instead of outside DRAM. It classifies 224x224 RGB image inputs at more than 140fps with peak power consumption at less than 300mW and an accuracy comparable to the VGG benchmark. The CNN-DSA accelerator is reconfigurable to support CNN model coefficients of various layer sizes and layer types, including convolution, depth-wise convolution, short-cut connections, max pooling, and ReLU. Furthermore, in order to better support real-world deployment for various application scenarios, especially with low-end mobile and embedded platforms and MCUs (Microcontroller Units), we also designed algorithms to fully utilize the CNN-DSA accelerator efficiently by reducing the dependency on external accelerator computation resources, including implementation of Fully-Connected (FC) layers within the accelerator and compression of extracted features from the CNN-DSA accelerator. Live demos with our CNN-DSA accelerator on mobile and embedded systems show its capabilities to be widely and practically applied in the real world.
\end{abstract}
%%%%%%%%% BODY TEXT
\section{Introduction}
Computers have been divided into servers, desktop computers and embedded
computers\cite{hennessy2011computer}. The basic building blocks for application domain-specific integrated computers
include the input, output, data path, memory and control - as with an ordinary computer\cite{david2005computer}.
Typical performance metrics of the computer system include the execution time and power consumption\cite{hennessy2011computer}. 
Domain Specific Architectures (DSA) are the only path forward for improved performance and energy efficiency given the end of Moore’s Law and Dennard scaling\cite{hennessy2017computer}.

Cellular Neural Networks or Cellular Nonlinear Networks have
been applied to many different fields and problems including, but not limited to, image
processing since 1988\cite{chua1988cellular}. However, most of the prior art approaches are either based on software solutions (e.g., Convolutional Neural Networks (CNN)\cite{lecun1998gradient,krizhevsky2012imagenet,zeiler2014visualizing,VGG,ResNet}, Recurrent Neural Networks\cite{mikolov2010recurrent}, etc.) or based on hardware that are designed for other purposes (e.g., graphic processing, general computation, etc.). As a result, prior CNN approaches are too slow in term of computational speed and/or too, expensive, thereby impractical for processing large amounts of imagery data. The imagery data can be from any two-dimensional signals (e.g., a still photo, a picture, a frame of a video stream, etc.)

There has been much interest in designing ASIC (Application-Specific Integrated Circuit) chips for computer vision tasks to efficiently accelerate CNN. Chen, Krishna, and et al. designed and made 65nm Eyeriss with the peak performance at 42GOPS at 200MHz core clock and 60MHz link clock, resulting in a frame rate of 34.7fps on the five
convolutional layers in AlexNet and a measured power of 278mW at 1V\cite{chen2017eyeriss}. The power efficiency of Eyeriss equals 151GOPS/Watt.
Han, Liu, and et al. designed and made EIE with a processing power of 102 GOPS working directly on
a compressed network, corresponding to 3 TOPS on an
uncompressed network, and processes FC layers of AlexNet with a power dissipation of only 600mW\cite{han2016eie}. The power efficiency of EIE equals 170GOPS/Watt, or corresponding to 5TOPS/Watt on an uncompressed network. 
Chen, Du, and et al. designed and made 65nm DianNao capable of performing 452 GOP/s at 485mW\cite{chen2014diannao}. The power efficiency of Diannao equals 932GOPS/Watt.
Du, Fasthuber and et al. designed and made a 65 nm CMOS technology with a peak performance of 194 GOP/s at 320.10 mW\cite{du2015shidiannao}. The power efficiency of ShiDianNao equals  606GOPS/Watt. Chen, Luo and et al. designed and made 28nm 15.97Watt 5.58 TeraOps/s for 16-bit operation\cite{chen2014dadiannao}. The power efficiency of DaDianNao equals 349.4GOPS/Watt. Jouppi, Young and et al. designed and made TPU with 92TOPS typically uses 40 Watts\cite{jouppi2017datacenter}. The power efficiency of TPU equals 2.3TOPS/Watt.

We designed a Convolutional Neural Networks Domain Specific Architecture (CNN-DSA) accelerator for extracting features out of an input image. It processes 224x224RGB image at 140fps with ultra power-efficiency of 9.3TOPS/Watt and peak power less than 300mW. The die size of CNN-DSA accelerator is 7mm by 7mm. This architecture mainly focuses on inference, not training. 
The designed CNN-DSA contains a plurality of identical CNN
processing engines operatively coupled to input/output (I/O) data bus. A
CNN processing engine controller is configured on the IC for controlling various
operations of the CNN processing engines. Each CNN processing engine includes a
CNN processing block, a first set of memory buffers for storing imagery data and a
second set of memory buffers for storing filter coefficients.

In the sections below, we will first provide our hardware design for CNN-DSA accelerator, and then the hardware performance metrics measured on test bench, followed by its implementation on mobile and embedded platforms with live demos. After that, we will show that CNN-DSA can support majority of CNN layer types through algorithm designs on the accelerator, and proved that ResNet\cite{ResNet}, MobileNet\cite{howard2017mobilenets} and ShiftNet\cite{wu2017shift} are special cases of VGG\cite{VGG}-Type model. For convenience of this paper, we define the CNN model as VGG-Type CNN models, if the layer types only consist of ReLU, 3x3 convolution, and max pooling. Lastly, we design algorithms and show the experimental results on the CNN-DSA accelerator for real-world applications on mobile and embeded platforms, and even very low-end MCUs (Microcontroller Unit). The target is to reduce the computation outside the CNN-DSA accelerator by fully utilizing our low power and ultra efficient CNN-DSA accelerator, including implementation of FC (Fully Connected) layers inside CNN-DSA accelerator, and compression of features extracted from CNN-DSA accelerator. 

%-------------------------------------------------------------------------
\section{CNN-DSA hardware design}
\begin{figure}[t]
\begin{center}
%\fbox{\rule{0pt}{2in} \rule{0.99\linewidth}{0pt}}
   \includegraphics[width=0.9\linewidth]{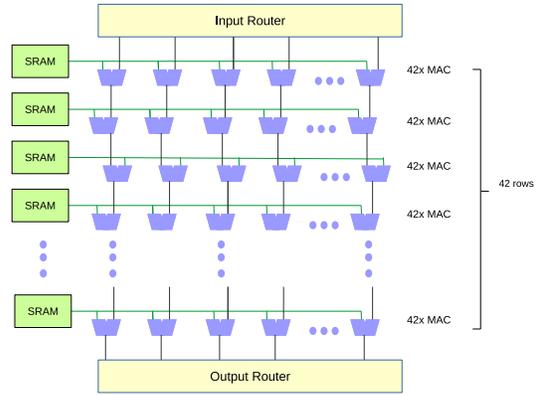}
\end{center}
   \caption{CNN-DSA processing engine}
\label{Architecture_design}
\end{figure}

For convolution on 2-D image, both the convolution filters and input data in each channel are two dimensional.
We design a matrix architecture of CNN-DSA Engine (CE)\cite{GTIpatent} shown as in Figure \ref{Architecture_design}.
CNN processing engines extract features out of an input image by performing multiple layers of 3x3 convolutions with
rectifications and 2x2 pooling operations. It includes a CNN processing block, a first set of memory buffers for
storing imagery data and a second set of memory buffers for storing filter coefficients.
The CNN processing block is configured to simultaneously perform 3x3 convolutions at
MxM pixel locations using received imagery data and corresponding filter coefficients, In Figure \ref{Architecture_design} we show our implementation of $M=14$.
Imagery data represents a ($M+2$)-pixel by ($M+2$)-pixel region of the input image. 
The CNN processing block further performs rectification and/or 2x2 pooling operations as
directed. When two or more CNN processing engines are configured on the integrated circuit (IC), the CNN
processing engines connect to one another as a loop via a clock-skew circuit for
cyclic data access. 

\begin{figure}[t]
\begin{center}
%\fbox{\rule{0pt}{2in} \rule{0.99\linewidth}{0pt}}
   \includegraphics[width=0.9\linewidth]{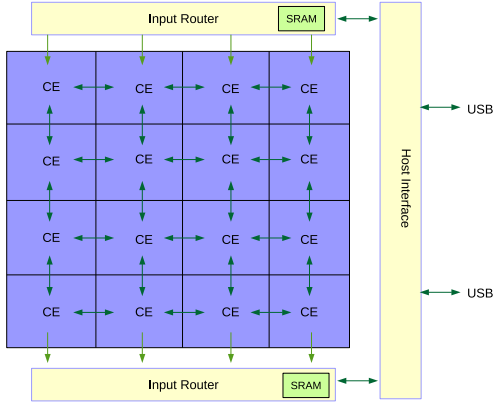}
\end{center}
   \caption{CNN-DSA block diagram}
\label{BlockDiagram}
\end{figure}

In Figure \ref{BlockDiagram} we show the block diagram of our CNN-DSA which
contains a plurality of identical processing engines operatively coupled to input/output (I/O) data bus.
The CNN-DSA contains $NE$ number of CNN processing engines connected in a loop via a
clock-skew circuit. In Figure \ref{BlockDiagram} we show our implementation of $NE=16$.
The CNN processing engines are connected to one
another to form a cyclic data access loop via a clock-skew circuit. The clock-skew
circuit enables the CNN processing engine receiving imagery data from a first neighbor
CNN processing engine while sending its own imagery data to a second neighbor CNN
processing engine. A CNN processing engine controller is also configured on the IC for controlling operations of the CNN processing engine.
The imagery data and filter coefficients are arranged in a specific scheme to fit the data
access pattern that the CNN based digital integrated circuit requires to operate. The
specific scheme is determined based on the number of imagery data, the number of
filters and the characteristics of the CNN based digital IC, such as the number of CNN processing engines, the connection direction
of clock-skew circuit and the number of the I/O data bus.
The method for arranging imagery data and
filter coefficients includes the following
steps and actions:
(a) determining number of imagery data groups required for storing $NIM$ sets of
imagery data in the CNN processing engines, with each imagery data group
containing $NE$ sets of the $NIM$ sets of imagery data, where $NE$ is the number of CNN processing engines connected in a loop via a
clock-skew circuit, and $NIM$ is a positive integer;
(b) circularly storing the $NE$ sets of the imagery data of each imagery data group
in the respective CNN processing engines;
(c) repeating (b) for the remaining imagery data groups;
(d) determining number of filter groups required for storing all filter coefficients
for $NF$ number of filters in the CNN processing engines, each filter group
containing $NE$ sets of filter coefficients and said each filter group being 
further divided into one or more subgroups with each subgroup containing a
portion of the $NE$ sets that correlates to a corresponding group of the
imagery data groups, where $NF$ is a positive integer;
(e) storing the portion of the $NE$ sets of filter coefficients in a corresponding one
of the CNN processing engines, the portion of filter coefficients being
arranged in a cyclic order for accommodating convolution operations with
imagery data received from an upstream neighbor CNN processing engine;
and
(f) repeating (e) for the remaining subgroups; and(g) repeating (e) and (f) for the
remaining filter groups. 

The CNN-DSA contains 16x42x42=28224 MAC units, and 9MB SRAM to hold the coefficients. All processings are in memory by CNN Domain Specific Floating Point (CNN-DSFP), enabling it to operate without external DRAM.
CNN-DSFP has totally 9-bits total for image data and activations, where 5 bits for mantissa and 4 bits for exponents. Filter coefficients are allocated 15-bits total, the mantissa 12 bits, exponents 2 bits, and 1 bit sign. 

Our design of CNN-DSA accelerator is optimized for VGG-Type CNN models. Specifically, it only consists of convolution kernels of 3x3 size. In the next section, we will prove that our design can implement majority of CNN models, including but not limited to, Resnet, MobileNet and ShiftNet.

\subsection{Hardware system implementation with CNN-DSA as coprocessor}
When building a hardware system, the CNN-DSA accelerator serves as a coprocessor. The host processor could be a mobile or embeded processor, or other host CPU,  which serves as controller. The interface between the host processor and CNN-DSA accelorator is USB or EMMC. To run inference with CNN-DSA accelerator, the host processor first loads weights and instructions of CNN model configurations to the on-chip SRAM. After that, the host processor sends image data to the CNN-DSA accelerator, and the data goes through the convolution processed by the coefficients loaded in the CNN-DSA accelerator. When the CNN-DSA accelerator finishes convolution, it signals the host processor and sends back the convolution output to host processor for next step processing. 

\section{CNN-DSA hardware performance measurments}
\subsection{Speed and power consumption measurement}
\begin{figure}[t]
\begin{center}
%\fbox{\rule{0pt}{2in} \rule{0.99\linewidth}{0pt}}
   \includegraphics[width=0.9\linewidth]{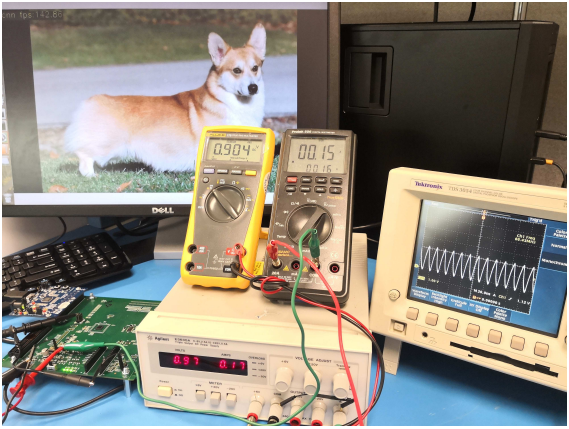}
\end{center}
   \caption{CNN-DSA accelerator speed and power consumption measurement. It is recommanded to zoom in this figure to see processing speed in fps at the top-left of the photo.}
\label{PowerMeasurement}
\end{figure}

The CNN-DSA accelerator speed and power consumption measurement is shown in Figure~\ref{PowerMeasurement}. The equipment used include:
A test board with CNN-DSA accelerator, a Fluke 179 DMM core voltage monitor, a Protek 506 DMM current monitor in Amps, an Agilent E3630A DC Power Supply which supplies core voltage power at 0.90 Volts normal, a Tektronix TDS 3054 Digital Phosphor Oscilloscope, a FTDI USB bridge UMFT601x, and Dell PowerEdge T30 PC with Ubuntu.

Figure~\ref{PowerMeasurement} shows that, we process the RGB image of size 224x224 at speed of 142.86fps when the CNN-DSA accelerator is set at 66MHz. Based on the readings from meters, the voltage is 0.904 volts, and the current is 0.15 Amps. So the power consumption is 0.904x0.15=0.1356W. Considering the different temperature conditions, we estimate that the power consumption in worst temprature conditions should be under 400mW, so the corresponding power efficiency is 28224x2x66M/0.4 = 9.3 TOPS/Watt. We also measured with same settings except that the frequency is 50MHz, and the power consumption is less than 300mW.
%~\ref{fig:onecol}

\subsection{Accuracy on ImageNet\cite{ILSVRC15} dataset}

Table \ref{ImageNet_benchmark} compares the accuracy and model size of the CNN models in our accelerator with other published models. For fair comparison, we only compared the convolutional layers sizes, without FC layers.
Here Gnet-1 is our compressed model using CNN-DSFP with same 13 layer convolutional architecture as VGG, and the FC layers are deployed outside the accelerator. In Section 6, we will show that FC can be implemented within our CNN-DSA accelerator as well.
Gnet-2 is almost the same architecture as Gnet-1 except that we halved the number of channels for layers 1 through 10, but kept the same number of 512 channels for layer 11 through 13 .
We can see that Gnet-1 compressed the original VGG model by more than 10x with the cost of only 4.5\% decrease of Top-1 accuracy.
And Gnet-2 further compressed the VGG model by more than 20x and the accuracy is still 1\% better than AlexNet.

\begin{table}
\begin{center}
\begin{tabular}{|c|c|c|}
\hline
Model & Accuracy Top-1 & Size(MB) \\
\hline%\hline
AlexNet\cite{krizhevsky2012imagenet} & 57.1 & 15.0 \\
VGG-16\cite{VGG} & \bf{71.5} & 58.9 \\
Googlenet\cite{szegedy2015going} & 69.8  & 23.2 \\
ShuffleNet-2x\cite{zhang2017shufflenet} & 70.9 & 19.2\\
1.0 MobileNet-224\cite{howard2017mobilenets} & 70.6  & 15.9\\
Compact DNN\cite{wu2017compact} & 68.9  & 13.6\\
Gnet-1(ours) & 67.0  & 5.5\\
Gnet-2(ours) & 58.1  & \bf{2.8}\\
\hline
\end{tabular}
\end{center}
\caption{Comparison of model size and accuracy on ImageNet\cite{ILSVRC15}.}
\label{ImageNet_benchmark}
\end{table}

\section{CNN-DSA accelerator deployed on mobile and embedded platforms}
\begin{figure}[t]
\begin{center}
%\fbox{\rule{0pt}{2in} \rule{0.99\linewidth}{0pt}}
   \includegraphics[width=0.9\linewidth]{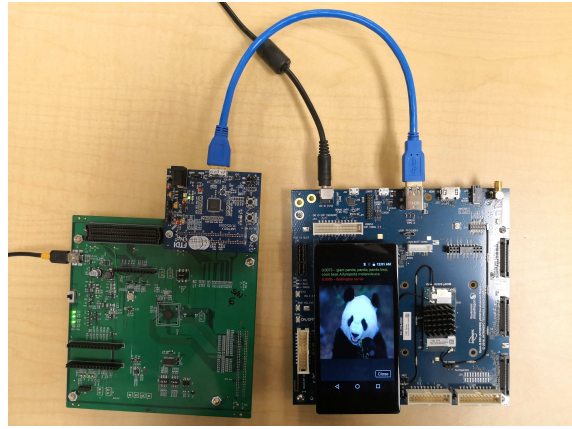}
\end{center}
   \caption{Image recognition on CNN-DSA accelerator connected to mobile platform.}
\label{MobilePlatform}
\end{figure}

\begin{figure}[t]
\begin{center}
%\fbox{\rule{0pt}{2in} \rule{0.99\linewidth}{0pt}}
   \includegraphics[width=0.9\linewidth]{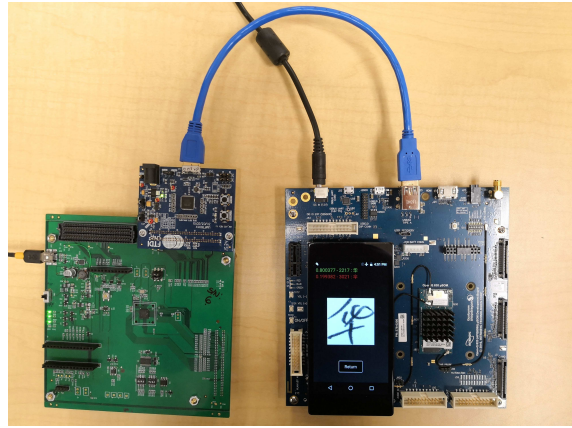}
\end{center}
   \caption{OCR on CNN-DSA accelerator connected to mobile platform.}
\label{MobileOCR}
\end{figure}

Figure \ref{MobilePlatform} shows an image classification demo on a mobile platform. We use a Qualcomm 820 development board with Android system. It has 8 cores, with 4 at 1.6GHz and 4 at 2.1GHz. The memory is 2.8GB and storage is 19GB. The interface to CNN-DSA accelerator is USB 3.0.
Figure \ref{MobileOCR} shows Chinese handwritten OCR (Optical Character Recognition) demo on the same embedded platform.

\section{Algorithm design on CNN-DSA to support various CNN models}
In this section, we design algorithms for CNN-DSA to implement various CNN models. And we prove that the majority of CNN layer types can be replaced and implemented by VGG-Type layers with 3x3 convolutional filters.
\subsection{ResNet is a special case of VGG-Type model\cite{GTIpatentResnet}}
\begin{figure*}
\begin{center}
%\fbox{\rule{0pt}{2in} \rule{1.1\linewidth}{0pt}}
   \includegraphics[width=0.6\linewidth]{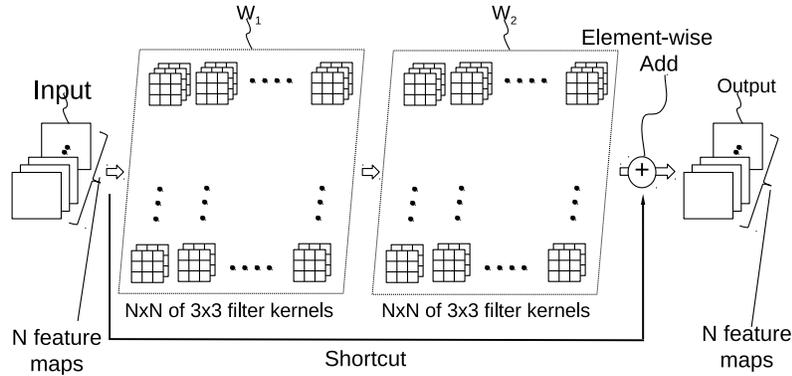}
\end{center}
   \caption{Example of a shortcut layer.}
\label{ResNet_original}
\end{figure*}

\begin{figure*}
\begin{center}
%\fbox{\rule{0pt}{2in} \rule{1.1\linewidth}{0pt}}
   \includegraphics[width=0.99\linewidth]{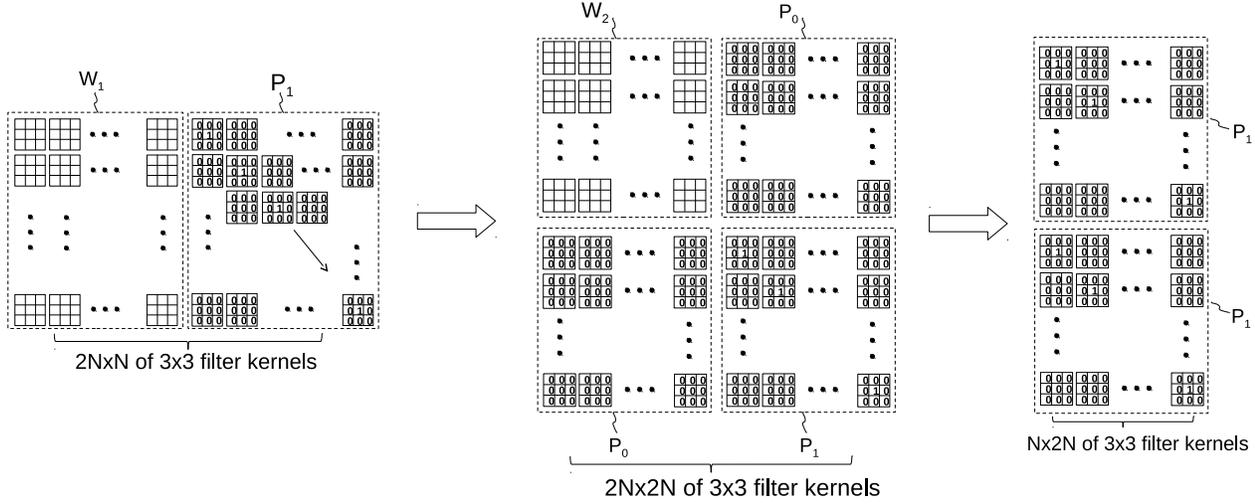}
\end{center}
   \caption{Implement Resnet in VGG-Type 3x3 convolutional filters.}
\label{Wholeblock_Resnet_in_VGG}
\end{figure*}

In addition to convolutional layers, activation layers and pooling layers, ResNet \cite{ResNet} requires shortcut layers. It typically contains an original convolutional layer W1 along with a second original convolutional layer W2 followed by element-wise add operations, as shown in Figure \ref{ResNet_original}. 
Because 3x3 convolutional operations are conducted with very fast speed in the CNN-DSA accelerator, it would therefore be desirable to implement deep neural network using 3x3 convolutional filter kernels to replace such operations in a CNN-DSA. By doing a network surgery and modifying the architecture of the trained ResNet, we can replace shortcut layer in Figure \ref{ResNet_original} by a set of three particular convolutional layers of multiple 3x3 filters as shown in \ref{Wholeblock_Resnet_in_VGG}. 
In Figure \ref{Wholeblock_Resnet_in_VGG}, first particular convolutional layer, which contains 2NxN of 3x3 filter kernels formed by placing original NxN of 3x3 filter
kernels of W1 in the left side and NxN of 3x3 filter kernels of an impulse response (or, identity-value) convolutional layer P1 in the right side. Each of the 3x3 kernels of P1 contains numerical value "0" except for the kernels located on the diagonal of the NxN kernels. Each of the diagonal kernels contains numerical value "0" in each of the eight perimeter positions and "1" in the center position. All off-diagonal kernels contains nine "0". Because of this, the first particular convolutional layer is configured for N-channels or N-'feature maps' input with 2N-channels output.
The second particular convolutional layer contains 2Nx2N of 3x3 filter kernels formed by placing NxN of 3x3 filter kernels of the second original convolutional layer W2 in the upper left corner and NxN of 3x3 filter kernels of impulse response convolutional layer P1 in the lower right corner, and two zero-value convolutional layers P0 in either off diagonal corner. The zero-value convolutional layers P0 contains NxN of 3x3 filter kernels with all zero numerical values in each of the 3x3 kernels. As a result, the second particular convolutional layer is configured for 2N-channel input and 2N-channel output. 
The third replacement convolutional layer contains Nx2N of 3x3 filter kernels formed by two impulse response convolutional layer P1 each containing NxN of 3x3 filter kernels in a vertical stack. As a result, the third particular convolutional layer is configured for 2N-channel input and N-channel output.
It is trivial to see that the output from Figure \ref{Wholeblock_Resnet_in_VGG} is equal to that from Figure \ref{ResNet_original}. By repeating the above mentioned method several times for all the shortcut in Resnet, we can convert all the shortcut into VGG type layers. So, ResNet is a special case of VGG-Type network. One point to mention here: we can only deploy shallow or narrow ResNet due to the memory limitation of the accelerator. 
\subsection{MobileNet is also special case of VGG-Type model\cite{GTIpatentMobileNet}}
\begin{figure}
\begin{center}
%\fbox{\rule{0pt}{2in} \rule{1.1\linewidth}{0pt}}
   \includegraphics[width=0.99\linewidth]{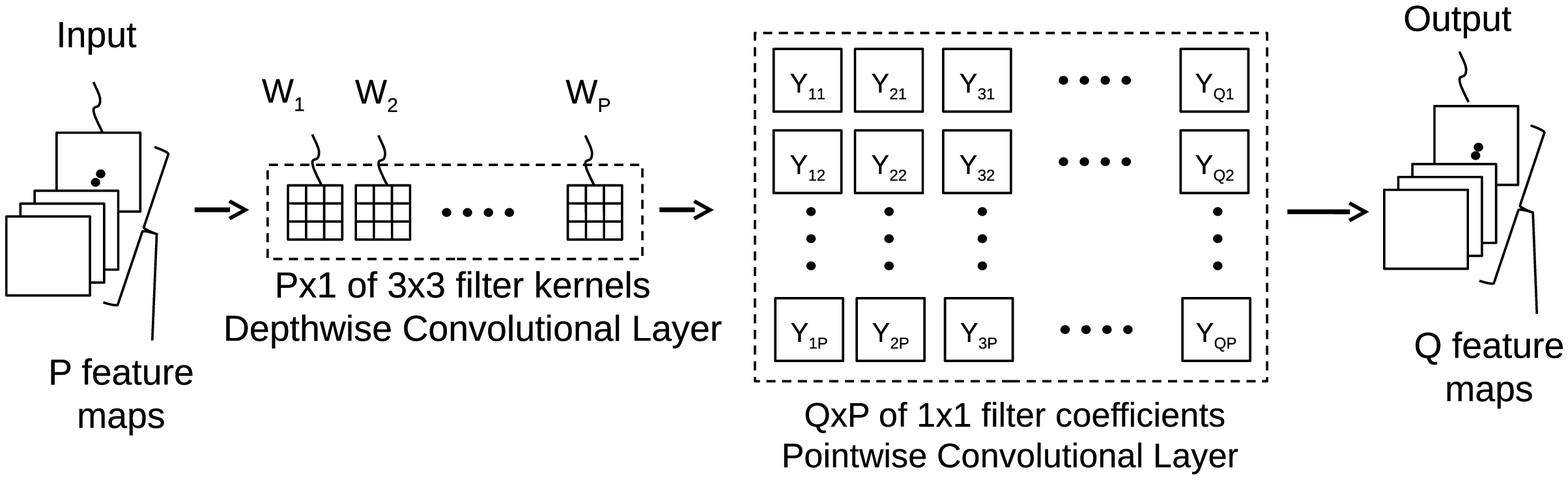}
\end{center}
   \caption{Depth-wise convolution followed by point-wise convolution.}
\label{MobileNet_Original}
\end{figure}

\begin{figure}
\begin{center}
%\fbox{\rule{0pt}{2in} \rule{1.1\linewidth}{0pt}}
    \includegraphics[width=0.99\linewidth]{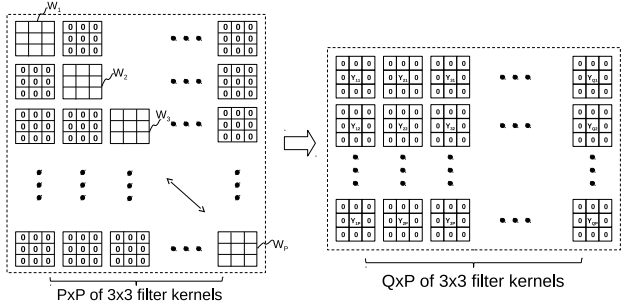}
\end{center}
   \caption{Implement MobileNet with VGG-Type 3x3 convolutional filters.}
\label{MobileNet_in_VGG_Type}
\end{figure}

In additional to convolutional layers, activation layers and pooling layers, MobileNet\cite{howard2017mobilenets} requires operations of depthwise separable layers. It typically contains a combination of a depthwise convolutional layer followed by a pointwise convolutional layer as shown in Figure \ref{MobileNet_Original}. Input and depthwise convolution contain P feature maps or channels, while the output contains Q feature maps. P and Q are positive integers. Pointwise convolutional layer contains QxP of 1x1 filter coefficients.
Similarly, we change the architecture of the trained MobileNet by replacing the depthwise convolution with the combination of multiple 3x3 filters, and set the weight to a specific value as shown in Figure \ref{MobileNet_in_VGG_Type}, such that these filters are equal to the functionality of a depthwise convolution. 
Figure \ref{MobileNet_in_VGG_Type} shows an example of the first replacement convolutional layer, which contains PxP of 3x3 filter kernels formed by placing Px1 of 3x3 filter kernels (W1, W2, ..., WP) of the depthwise convolutional layer on respective diagonal locations. The remaining off-diagonal locations of the PxP of 3x3 filter kernels are filled by zero-value 3x3 filter kernels. As a result, the first replacement convolutional layer is configured for replacing the depthwise convolutional layer. 
The second replacement convolutional layer shown in Figure \ref{MobileNet_in_VGG_Type} contains QxP of 3x3 filter kernels formed by placing the QxP of 1x1 filter coefficients Y11, Y21, Y31, ..., YQ1, Y12, Y22, ..., YQ2, ..., Y1P, Y2P, ..., YQP of the pointwise convolutional layer in the center position of the respective QxP of 3x3 filter kernels, and numerical value zero in eight perimeter positions. As a result, the second replacement convolutional layer is configured for replacing the pointwise convolutional layer.
Similarly, we utilize many 3x3 filters to implement the equivalent depth wise convolution in the deep neural network, and finally replace the depthwise convolution by the VGG-Type layers. Using the same scheme, Inception\cite{szegedy2015going} and ShiftNet \cite{wu2017shift} can also be proved as special cases of the VGG-Type model.
\section{Algorithm design and experiments on CNN-DSA for various application scenarios}
Considering CNN-DSA accelerator applications in real-world scenarios, we design algorithms that best utilize the fast speed and low power consumption of CNN-DSA accelerator, in order to best fit mobile and embedded platforms.
\subsection{Basic model applications with CNN-DSA accelerator and experiment result on CASIA-HWDB\cite{CASIA2013}}
Traditional deep neural network architectures for 2-D inputs generally have two parts: convolution layers and FC layers. Notably, convolution layers require less storage for holding coefficients but require significant amounts of computation (for example, VGG16 \cite{VGG} requires 15 TFLOPs) for Mult-Adds due to the repeated applications of filters. On the contrary, FC layers require less computation for Mult-Adds but necessitates a significant amount of storage (for example, VGG16 requires storage for 123M FC coefficients) for storing coefficients due to inner-products.

VGG-Type models are optimally accelerated in CNN-DSA. The basic application mode is to use the CNN-DSA accelerator as the feature extractor, so the most heavy-lifting convolutional computation is taken care of by the power efficient accelerator. The output features from the CNN-DSA accelerator connect to the outside-accelerator devices for the FC computation. The FC which is typically handled by cpu or gpu, consumes more storage and memory but little computation and power.
We show one experiment for this basic application mode on the CNN-DSA accelerator. The data set we used here is the offline handwritten isolated character recognition CASIA-HWDB\cite{CASIA2013} released by CASIA (the Institute of Automation of Chinese Academy of Sciences (CASIA)). We used the designed training and testing data set of unconstrained offline Chinese handwriting database level-1 set of GB2312-80 with 3755 Chinese characters to train our model and evaluate accuracy performance. We use the same architecture as VGG-16\cite{VGG} and trained CNN-DSFP fixed point model to load it into our CNN-DSA accelerator, and the fully connected is processed by mobile device with Android system, as shown in Figure~\ref{MobileOCR}. We got 94.95\% accuracy on the testing data running low power CNN-DSA accelerator connected to a mobile device. The ICDAR\cite{CASIA2013} competition on this data set gave 94.77\% in 2013 \cite{CASIA2013} and 92.18\% in 2011 \cite{CASIA2011}.

\subsection{Mobile friendly: compress extracted features directly from CNN-DSA accelerator\cite{GTIpatentDirectlyConv}}
For tasks like object classification and face authentication, feature extraction from input is useful. Traditional methods of face feature extraction use one or several layers of FC layers in the CNN network. Through inner product computation, the output of FC layer dimension would be reduced from convolution layers of very high dimensional data. For example, the convolution output after pooling in VGG-16 gives a 512x7x7=25088 high dimensional feature which the FC layer will project into a relatively low dimensional space, like 4096, or 1024, or 128, as done by\cite{VGGface}. But the disadvantage of this feature extraction is that the model size remains very large, especially for mobile devices. For example, the FC layer connecting convolution layers in VGG-16 will require 25k*4k=100M parameters. At the same time, the runtime performance will be low due to the high computation complexity. 

Our CNN-DSA accelerator can load the entirety of the VGG net into the chip, and completes the heavy-lifting convolution computations at local. We design a method of extracting features directly from convolution layers in the CNN-DSA with very small number of channels. For a given image input, the output from CNN-DSA accelerator completes feature extraction without using FC layers. After our invention method is implemented in our CNN-DSA accelerator, it requires minimum hardware support for feature extraction. Feature vectors are obtained directly from the convolution layer output, instead of from the FC. We also use very small number of channels from the convolution layer. Thus, the output from CNN-DSA accelerator requires minimum hardware requirement from outside the accelerator.

We show our method implemented in CNN-DSA in a face authentication experiment. We have a collected data set with 21278 different people, each with at least 10 pictures. First, we trained our model using same model architecture as VGG using our CNN-DSFP fixed point. Then we changed the number of channels for 5\_3 from 512 to 8,4,2,1, but still connected to the same FC layers of fc6 of size 4096, fc7 of size 4096, and fc8 of size 21278. After training, the convolutional layers of the model are loaded into CNN-DSA, and can be used directly for face feature extraction from output of convolutional layer 5\_3. The FC layers are no longer used. Input face image will be feature-extracted directly from the output of CNN-DSA. Thus, the extracted feature will be only a vector of 49x1 (7x7x1) for the model with 5\_3 set to 1 channel. This low dimensional data is feasible for computation in very low end devices like MCUs (Microcontroller Unit).

We use the LFW \cite{LFWTech} data set for face authentication performance evaluation. For preprocessing before performance evaluation, we only made a cropping using dlib\cite{dlib09} face detection on original LFW data set. The same preprocessing used in VGG Face \cite{VGGface} gives an accuracy of 92.83\% with no alignment and no embedding. Table \ref{CompressChannels} shows the result on LFW unrestricted setting with compressed number of channels from CNN-DSA. 
\begin{table}
\begin{center}
\begin{tabular}{|c|c|}
\hline
\#Channels & LFW Verification Accuracy\\
\hline%\hline
512 & \bf{95.57} \\
8 & 94.92 \\
4 & 94.76\\
2 & 94.42\\
\bf{1} & 94.25\\
\hline
\end{tabular}
\end{center}
\caption{Compress the number of channels of VGG 5-3 layer}
\label{CompressChannels}
\end{table}
We have decreased the number of channels all the way down from 512 to 1 but accuracy performance degradation is negligible. In this way we compressed the size of extracted features. The trade off of using this method is that the robustness for face verification degrades a little bit. But it minimizes the computation in outside the accelerator devices. 

\subsection{Implement FC inside CNN-DSA accelerator\cite{GTIpatentFCinChip}}
FC computations are mainly composed of inner-products, which can easily be replaced by convolution layers if the input data size of the convolution layers is the same as that of the inner-product. However, our CNN-DSA accelerator only supports 3x3 convolution kernel, because large kernel size convolutional layers are not efficient in matrix products. As pointed out by the VGG paper, two connected 3x3 convolution kernels have the same reception field as that of a 5x5 kernel, and three connected 3x3 kernels can have the same reception field as that of 7x7 kernel.
With operations of convolutional layers performed very fast in our CNN-DSA accelerator, the computation bottleneck in deep learning networks is in FC which are processed by CPU or GPU. We provide a solution for utilizing CNN-DSA accelerator in classification tasks without FC. We use multiple layers of 3x3 kernels to approximate larger-sized kernels. By using L layers of connected 3x3 kernels, we can approximate the inner-product for an input feature map of (2*L+1)*(2*L+1) size. For example, for a feature map input of size 7x7, we can use L=3 layers of connected 3x3 convolution kernels to approximate the inner-product. For a feature map input of size 7x7, applying a 3x3 convolution without padding will output a 5x5 feature map. Applying a second layer of size 3x3 without padding, connected to this 5x5 feature map, will result in an output of a 3x3 feature map. Finally, applying a third layer of size 3x3 without padding, connected to this 3x3 feature map, will output a 1x1 map, which is a scalar. This scalar can be treated as equal to the result given by an inner-product of the original 7x7 feature map. We can still use L$'$ $>$ (F-1) /2 (L$'$ is another possible number of layers of connected 3x3 kernels) to arrive at the (2L$'$+1) layers approximation for the inner-product by simply increasing the redundancy. So, FC is implemented within the CNN-DSA accelerator.

\begin{figure}
\begin{center}
%\fbox{\rule{0pt}{2in} \rule{1.1\linewidth}{0pt}}
   \includegraphics[width=0.9\linewidth]{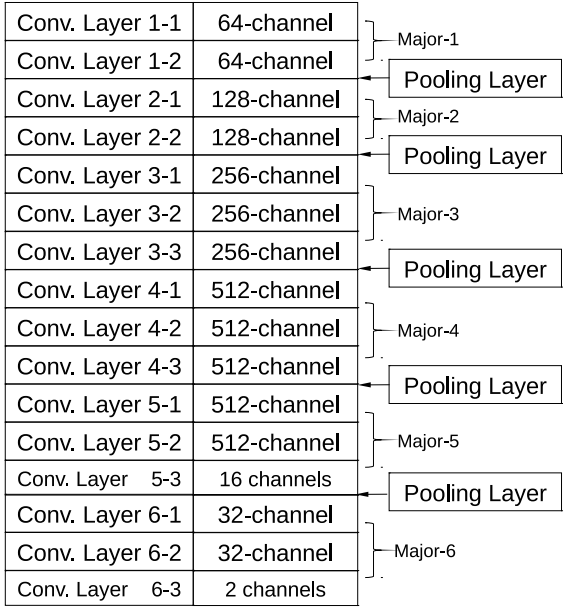}
\end{center}
   \caption{CNN model architecture of implementation of Fully Convolutional Classification using only 3x3 convolution kernels.}
\label{DirectlyUsingConvAsFeatureExtraction}
\end{figure}

We show an example of implementing FC in CNN-DSA on Kaggle competition cats and dogs data set\cite{asirra-a-captcha-that-exploits-interest-aligned-manual-image-categorization}. The CNN architecture is shown in \ref{DirectlyUsingConvAsFeatureExtraction}. The first five major layers are same as VGG-16, except that the 5\_3 layer has 16 channels. And we add a sixth major layer with 3 layers, 32 channels, 32 channels and 2 channels. On this data set, we achieved 97.4\% accuracy with the FC and convolution layers all inside the accelerator. This gives the same accuracy as the traditional (convolution + FC layered) solutions.This model is also using our CNN-DSFP fixed point model.

Although we only show a special case of binary classification, it indicates the potential of using our chip in cheap, low-end computation devices. Classification computation is done within the chip, while only a tiny portion of calculation is done outside the chip. 

%-------------------------------------------------------------------------
\section{Conclusion and future work}
In conclusion, we have designed a CNN-DSA accelerator which achieved a power consumption of less than 300mW and an ultra-power efficiency of 9.3TOPS/Watt. Demos on mobile and embedded systems show its applications in real-world scenarios. This 28nm Two-Dimensional CNN-DSA accelerator attains a 140fps at an accuracy comparable to that of the VGG architecture. The CNN-DSA accelerator has four main advantages: First, it is a CNN-specific matrix architecture and distributes memory blocks as the center of the processing unit. Second, it is data driven and enables full parallel processing. Third, CNN domain-specific floating point data structures efficiently use memory and completes all processing in internal memory rather than outside DRAM. Fourth, it allows for flexible process unit expansion. We also proved that shorcut layer types and depthwise separable convolution layer types are special cases of VGG-Type layers. The current CNN-DSA accelerator supports ResNet and MobileNet, but it is originally optimized for VGG-Type CNN models. Future work involves designing a CNN-DSA to support ResNet and MobileNet, among other architectures. 

{\small
\bibliographystyle{ieee}
\bibliography{egbib}
}

\end{document}